\documentclass{article}
\usepackage[a4paper, left=3cm, right=3cm, top=2cm, bottom=2cm]{geometry}
\usepackage{graphicx}
\usepackage{times}  
\usepackage{helvet} 
\usepackage{courier} 
\usepackage[hyphens]{url} 
\usepackage{graphicx} 
\usepackage{caption} 
\usepackage{algorithm}
\usepackage{algorithmic}
\usepackage[utf8]{inputenc} % allow utf-8 input
\usepackage[T1]{fontenc}    % use 8-bit T1 fonts
\usepackage{url}            % simple URL typesetting
\usepackage{booktabs}       % professional-quality tables
\usepackage{amsfonts}       % blackboard math symbols
\usepackage{nicefrac}       % compact symbols for 1/2, etc.
\usepackage{microtype}      % microtypography
\usepackage{xcolor}         % colors
\usepackage{bm}
\usepackage{times}
\usepackage{graphicx}
\usepackage{paralist,algorithmic,algorithm}
\usepackage{amsmath,mathrsfs,amssymb}
\usepackage{amsfonts}       % blackboard math symbols
\usepackage{nicefrac}       % compact symbols for 1/2, etc.
\usepackage{microtype}      % microtypography
\usepackage{booktabs}       % professional-quality tables
\usepackage{longtable}
\usepackage{threeparttable}
\usepackage{multirow}
\usepackage{multicol}
\usepackage{soul}
\usepackage{subfigure}
\usepackage[american]{babel}
\usepackage{algorithm}
\usepackage{algorithmic}
\usepackage{lipsum}
\usepackage{multicol}
\usepackage{url}
\usepackage{float}
\usepackage{subcaption}
\usepackage{makecell}   % 多行换行
\usepackage{tabularx}
\allowdisplaybreaks[3]
\usepackage{hyperref}

\usepackage{caption} % DO NOT CHANGE THIS AND DO NOT ADD ANY OPTIONS TO IT
\usepackage{authblk}

\title{Efficient Ternary Weight Embedding Model: \\Bridging Scalability and Performance}
\author[1]{\textbf{Jiayi Chen}\thanks{Lead author}}
\author[3]{Chen Wu}
\author[2]{Shaoqun Zhang\thanks{Corresponding author: \texttt{zhangsq@lamda.nju.edu.cn}}}
\author[3]{Nan Li}
\author[3]{Liangjie Zhang}
\author[3]{Qi Zhang}

\affil[1]{\centering School of Intelligent Science and Technology, Nanjing University, Suzhou, China\\
\texttt{jiaychen@smail.nju.edu.cn}
}
\affil[2]{\centering National Key Laboratory for Novel Software Technology, Nanjing University, Nanjing, China\\
\texttt{zhangsq@lamda.nju.edu.cn}
}
\affil[3]{\centering Microsoft AI, Beijing, China\\ 
\texttt{\{wu.chen, nan.li, liazha, zhang.qi\}@microsoft.com}
}

\date{}

\begin{document}

\maketitle

\begin{abstract}
Embedding models have become essential tools in both natural language processing and computer vision, enabling efficient semantic search, recommendation, clustering, and more. However, the high memory and computational demands of full-precision embeddings pose challenges for deployment in resource-constrained environments, such as real-time recommendation systems.
In this work, we propose a novel finetuning framework to ternary-weight embedding models, which reduces memory and computational overhead while maintaining high performance.
To apply ternarization to pre-trained embedding models, we introduce self-taught knowledge distillation to finalize the ternary-weights of the linear layers. 
With extensive experiments on public text and vision datasets, we demonstrated that without sacrificing effectiveness, the ternarized model consumes low memory usage and has low latency in the inference stage with great efficiency. 
In practical implementations, embedding models are typically integrated with Approximate Nearest Neighbor (ANN) search. 
Our experiments combining ternary embedding with ANN search yielded impressive improvement in both accuracy and computational efficiency. The repository is available at \href{https://github.com/dataparameters/Ternary-Embedding-Models}{here}.
\end{abstract}

\section{Introduction}\label{sec:introduction}

In recent years, embedding models have emerged as fundamental components in a wide array of applications within natural language processing (NLP) and computer vision (CV). 
The utility of embeddings spans across various tasks, including semantic search, recommendation systems, clustering, and classification, enabling machines to process and retrieve information with remarkable efficiency and accuracy.

Despite their widespread adoption and success, embedding models, particularly those based on large-scale architectures like BERT \cite{devlin:bert} and ViT \cite{dosovitskiy:vit}, face significant challenges related to memory consumption and computational overhead. 
The high precision of weights in these models (typically 32-bit floating-point numbers) results in substantial memory footprints, which hinders their deployment in resource-constrained environments such as mobile devices, real-time search or recommendation systems, and edge computing platforms. Moreover, the computational demands associated with full-precision embeddings can lead to increased latency, limiting the scalability and responsiveness of applications that rely on these models.

Model quantization has emerged as a promising solution to mitigate these challenges by reducing the precision of model weights, thereby decreasing memory usage and accelerating inference without markedly compromising performance. While post-training quantization (PTQ) \cite{jacob:ptq} and quantization-aware training (QAT) \cite{jacob:qat} have shown effectiveness in compressing models. 
Binary-weight networks, which constrain weights to \{-1, +1\}, have demonstrated the potential to eliminate multiplications in matrix operations, relying solely on additions. However, the binary constraint often leads to significant information loss, adversely affecting model performance \cite{rastegari:xnor-net}.

Ternary-weight networks, which extend binary quantization by introducing a zero state \{-1, 0, +1\}, present a balanced trade-off between compression and performance.
The inclusion of zero allows for sparsity in the weight matrices, akin to dropout regularization, which can enhance model robustness and maintain a higher degree of representational capacity compared to binary networks. Despite these advantages, the ternarization of pre-trained embedding models remains under-explored, particularly in the context of fine-tuning techniques that preserve the efficacy of full-precision models.
In addition, with ternary-weight networks, multiplication operations can also be replaced by addition operations.

In this work, we address the aforementioned challenges by introducing a novel finetuning framework that transforms pre-trained embedding models into ternary-weight networks. Our approach leverages self-taught knowledge distillation to calibrate the ternary weights of linear layers, ensuring that the compressed models retain high performance across diverse tasks. By adopting ternary weights, our models achieve significant reductions in memory usage and inference latency, making them well-suited for deployment in environments with resource constraints.

The contributions of our work are outlined as follows:
\begin{itemize}
        \item We are the first to apply ternary quantization to embedding models and the pre-trained language models through fine-tuning. This approach provide a more flexible and energy-efficient way for model reuse rather than reconstruction and retraining from scratch.
	\item We thoroughly evaluated our fine-tuning method on text (BERT-based) and image (ViT-based) embedding models using the MTEB benchmark and real-world image datasets. The ternary-weight models obtained demonstrates a comparable performance with 32-bit models while significantly reducing latency and storage.
	\item Recognizing that embedding models are often coupled with ANN search techniques in practical applications, we integrate our ternary-weight text embedding models with popular ANN algorithms. The results showcase not only improved retrieval speeds but also maintained or even enhanced accuracy, enabling the practical viability of our approach in real-time recommendation systems.
\end{itemize}

The remainder of the paper is organized as follows. 
Section~\ref{sec:related work} introduces the related work on model quantization and embedding models.
Section~\ref{sec:methodology} proposes the ternary-weight models and the self-distillation method.
Section~\ref{sec:experiments} conducts extensive experiments to evaluate the ternary-weight models obtained through our method. 
Section~\ref{sec:discussion} concludes and discusses the limitations and future work.

\section{Related Work}\label{sec:related work} 

\subsection{Quantization} 
Quantization aim at reducing the precision of a model's parameters from high-precision format like 32-bit floating-point to lower-precision formats, commonly 8-bit integers.
The reduction decreases the model's memory footprint and computational requirements, thereby enhancing inference speed and enabling deployment on resource-constrained devices without significantly compromising accuracy.
% Quantization of deep neural networks is commonly approached through two primary methods: 
Two primary approaches to quantization have been comprehensively studied:
\begin{enumerate} 
    \item \textit{Post-Training Quantization (PTQ)}: This method involves first training a model to convergence using full-precision weights. Subsequently, the weights are converted to lower precision without further training. PTQ is typically computationally inexpensive compared to training processes \cite{nagel2019data, banner2019post}.
    \item \textit{Quantization-Aware Training (QAT)}: In this approach, quantization is integrated during the initial training phase or during additional fine-tuning stages. QAT generally yields better performance than PTQ but requires additional computational resources and access to representative training data \cite{jacob2018quantization, krishnamoorthi2018quantizing}.
\end{enumerate}

\subsection{Embedding Models}\label{subsec:embedding_models} 
In both natural language processing (NLP) and computer vision (CV), embedding models have been foundational in enabling efficient processing and retrieval, as embeddings allow data similarity and relevance to be efficiently assessed through vector operations.
\begin{itemize}
    \item In NLP, embeddings are often derived from transformer-based language models like BERT \cite{devlin2018bert} or sentence encoders \cite{reimers2019sentence}. These embeddings capture semantic similarity between texts, enabling applications like search, recommendation, and clustering.
    \item In CV, embeddings are generated by models like ViT \cite{dosovitskiy2020image} that encode visual features into representations, which are useful for classification, object detection, and content-based image retrieval.
\end{itemize}

Given the memory and compute costs associated with embeddings generated from full-precision models, quantized embeddings offer a promising alternative \cite{gong2014compressing, stock2019bit}. 
By reducing the precision of embeddings (e.g., from 32-bit to ternary weight), storage requirements and computational costs can be significantly reduced. 
Our work focuses on developing ternary-weight embedding models for text (BERT-based) and image (ViT-based) applications, optimizing efficiency while maintaining comparable performance to their full-precision counterparts.

\section{Methodology}\label{sec:methodology}

\subsection{Ternary-weight Models}\label{subsec:model}
Recent advancements in large language models (LLMs) have been significant; however, their deployment and online servicing necessitate considerable energy consumption and additional resource expenditures. Consequently, accelerating model inference has become a critical area of research in practical applications, which allows tasks to be completed more rapidly without necessitating hardware upgrades.
Matrix multiplications, particularly those involving 32-bit floating-point operations, are computationally intensive. 
For instance, a single 32-bit floating-point multiplication consumes 1.31 pJ, while a single addition requires only 0.38 pJ on a 7nm processor~\cite{jouppi:energy7nm}. 
To reduce the computational load associated with these operations, employing binary or ternary weights is a viable alternative, as they can replace multiplication with simpler operations like addition and subtraction. 
In contrast, more complex quantization methods still depend on multiplication, maintaining higher computational demands.

Binarizing weights to values such as -1 and 1 is a prevalent method for extreme 1-bit network compression. 
This approach has shown promise in language models, as demonstrated by BitNet~\cite{wang:bitnet}. 
When integrated with appropriate computational kernels, binary-weight networks can replace multiplication operations in matrix computations with more efficient addition operations. 
However, this binarization process introduces significant information loss compared to full-precision weights. 
A critical issue arises when values near zero are forced to -1 or 1, potentially misrepresenting their original significance and disproportionately affecting the model's performance. 

% The most extreme 1-bit network compression typically uses binary weights with values of -1 and 1, and weight binarization has already made some progress in language models~\cite{wang:bitnet}. When paired with suitable computational kernels, binary-weight networks can eliminate costly multiplications in matrix operations, relying solely on additions. However, binary weights suffer from substantial information loss compared to full-precision weights. This is especially problematic because binarization can force values near 0 to be set to either 1 or -1, potentially misrepresenting their original significance and giving them undue importance relative to other values.

In this work, we explore the ternarization of weights. 
% Expanding the range of weight values to -1, 0, and 1 offers a better solution, representing the most complex option that still avoids multiplication in matrix calculations. 
This approach enhances representational capacity while maintaining computational efficiency by eliminating multiplication operations in matrix calculations. Ternary weights offer a balance between the simplicity of binary weights and the precision of full-precision weights, providing a more nuanced representation that can improve model performance without incurring significant computational overhead.

% Ternary weights add a 0 to the binary set in pratical, resulting in a bit width of $\textrm{log}_2 3$, or 1.58 bits per value. 
% We argue that the inclusion of 0 functions could be see as adding dropout~\cite{srivastava:dropout} to the network. In essence, dropout creates multiple variations of the network during training and ultimately combines them into a merge. Similarly, if we combine two binary-weight networks with values in \{-1, 1\}, the resulting weight values will be in \{-2, 0, 2\}. With appropriate scaling, this can be made equivalent to \{-1, 0, 1\}. Thus, a ternary-weight network could approximate the representational power of two combined binary-weight networks, without significantly increasing computational or storage requirements.
Ternary weight networks, which utilize weights of -1, 0, and 1, effectively represent each weight with approximately 1.58 bits, as $\textrm{log}_2 3$ is approximately equal to 1.58. The inclusion of zero-valued weights introduces sparsity, akin to the effect of dropout regularization. 
Dropout functions by randomly omitting units during training, thereby preventing over-fitting and enhancing generalization. Similarly, zero weights in ternary networks can be viewed as deactivating certain connections, promoting a form of implicit regularization.

Moreover, combining two binary-weight networks, each with weights in {-1, 1}, results in a weight set of {-2, 0, 2}. 
With appropriate scaling, this set can be transformed to {-1, 0, 1}, effectively forming a ternary-weight network. 
This demonstrates that a ternary-weight network can approximate the representational capacity of two combined binary-weight networks without significantly increasing computational or storage demands. By leveraging the advantages of both binary and ternary representations, ternary-weight networks offer a balanced approach to model compression and efficiency.

\subsection{Self-Distillation}\label{subsec:distillation}
Training a ternary-weight model from scratch has yielded significant successes~\cite{ma:1.58bitLLM,yuan:vit1.58b}. 
However, the specified model architectures need to be designed, and the training process undeniably requires extra resources.
A key concern is that abandoning mature and powerful full-precision models to train ternary models from scratch results in a significant waste of valuable resources. 
To achieve efficient and effective models with reduced computational demands, fine-tuning ternary-weight models from pre-trained full-precision counterparts presents a promising direction for exploration.

% \textbf{Knowledge distillation.} Knowledge distillation~\cite{hinton:teacher-student} provides a class of methods for deriving lightweight models from existing comprehensive models. 
% To compress model using one-hot encoding, the final classification results of complex models are typically not utilized to mitigate information loss. Instead, the probability distributions of all categories are conveyed to the lightweight model as `knowledge'. This method is particularly well-suited for embedding models that produce continuous vector representations. 

Knowledge distillation~\cite{hinton:teacher-student} encompasses a suite of techniques aimed at deriving compact models from more complex counterparts. 
In this paradigm, rather than relying solely on the hard labels (i.e., one-hot encoded outputs) of the teacher model, the soft probability distributions over all classes are transferred to the student model. 
This approach effectively conveys the nuanced information captured by the teacher, thereby mitigating potential information loss during compression. Such methodologies are particularly advantageous for embedding models that generate continuous vector representations, as they facilitate the preservation of intricate relational information inherent in the data.

\begin{figure}[t]
	\centering
	\includegraphics[width=1\textwidth]{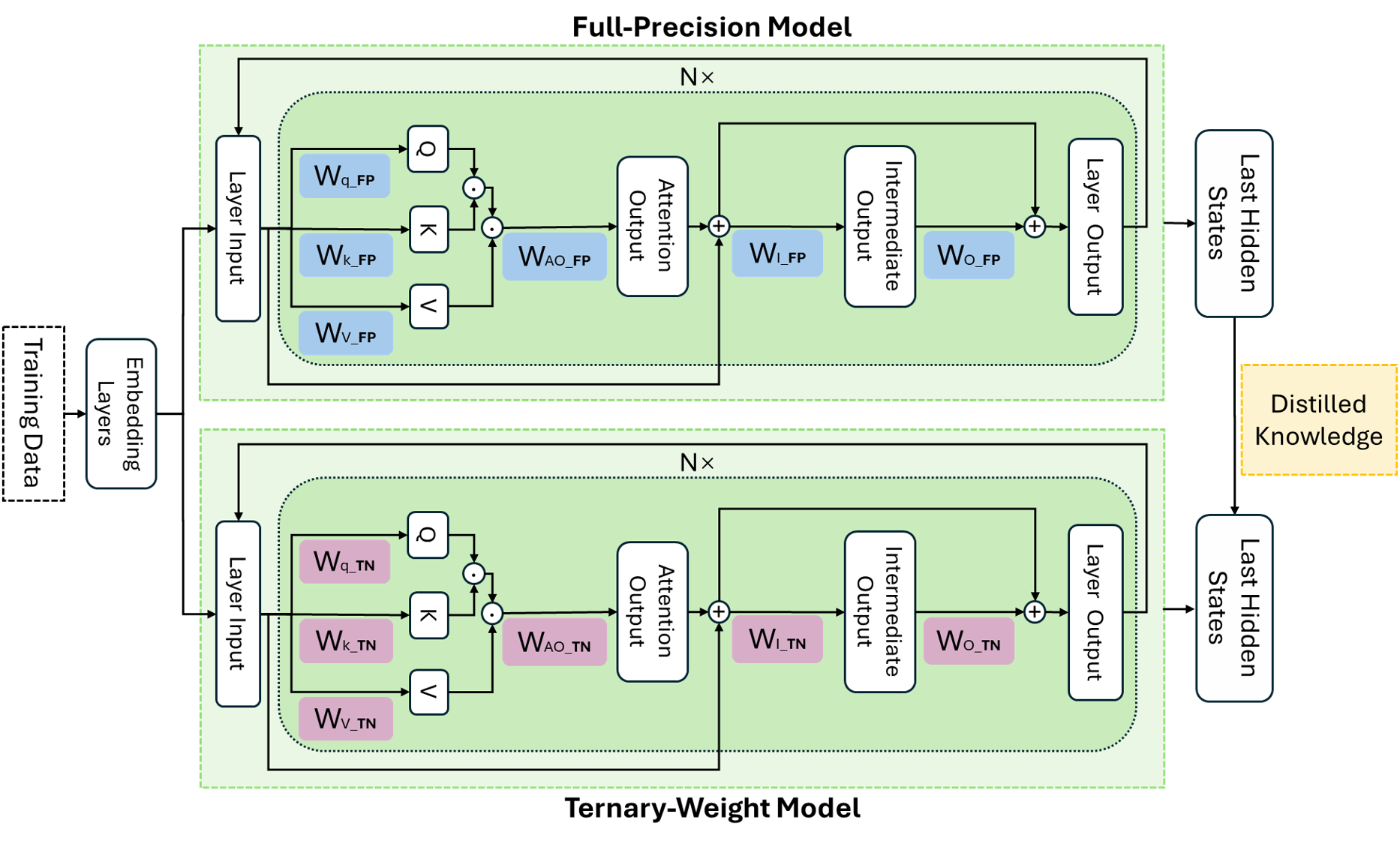}
	\caption{Fine-tuning flow. The ternary-weight model (bottom) is derived by replacing all linear layers of the full-precision model (top). The output of full-precision model acts as the target for the ternary-weight model.}
	\label{fig:trainflow}
\end{figure}

We introduce a self-distillation framework that maintains the original model's architecture by substituting only its linear layers.
The modified model serves as a draft model. By utilizing the outputs of a pre-trained, full-precision embedding model as the targets, we fine-tune the draft ternary-weight model using a limited dataset.
The training process is illustrated in Figure~\ref{fig:trainflow}.

\textbf{Forward and Backward Propagation.} Ternary Weight Network~\cite{li:TWN} is the earliest work on weight ternarization, which proposes an approach to achieve a ternary weight matrix $\tilde{\textbf{W}}$ by processing the full-precision weight matrix $\textbf{W}$. Given $\textbf{W}_{ij}$ as the element in the $i$-th row and $j$-th column of an $m\times n$ matrix, the optimal threshold for partitioning is $\gamma = 0.75\sum_{ij}|\textbf{W}_{ij}|/mn$. However, this requires the weights to follow the Gaussian distribution $\mathcal{N}(0,\sigma^2)$ with a standard deviation $\sigma$. 
In our work, we introduce a parameter $\beta$ to bypass this assumption. Assuming $\textbf{W}^l$ as the weight matrix of the $l$-th linear layer, the forward propagation with the layer input $\textbf{X}^{l}$ can be formulated as
\begin{equation}  \label{eq:forward}
\begin{aligned}
&\gamma^l=\frac{\beta}{mn}\sum_{ij}\left|\textbf{W}_{ij}^l\right| \ , \\
&\tilde{\textbf{W}}_{ij}^l = f\left(\textbf{W}^l\mid \gamma^l \right) = \left\{
\begin{aligned}
    1 \ , &\quad \textbf{W}_{ij}^l  > \gamma^l \ ,\\
    0 \ , &\quad-\gamma^l \leq \textbf{W}_{ij}^l \leq \gamma^l \ ,\\
    -1 \ , &\quad \textbf{W}_{ij}^l < -\gamma^l \ ,\\
\end{aligned}\right.\\
&\textbf{X}^{l+1}=\gamma^l\tilde{\textbf{W}}^l\textbf{X}^{l} \ ,
\end{aligned}
\end{equation}
where $\beta$ is empirically set (see Table~\ref{tbl:MTEB_xiaobu_different_beta} in Appendix~\ref{apd:beta}) and $f(\cdot)$ denotes the partition function. 
The derivative of $f(\textbf{W}\mid \gamma)$ is zero everywhere except at $\pm \gamma$, where it is non-differentiable. To enable backpropagation, we follow the traditional paradigm of Straight-Through Estimator~\cite{bengio:STE}, which is commonly used in training low-bit quantized models. Specifically, we directly set $\partial f\left(\textbf{W}\mid \gamma \right) / \partial \textbf{W}=1$ to estimate the gradient of the ternary weight $\tilde{\textbf{W}}$ with respect to $\textbf{W}$.

\section{Experiments}\label{sec:experiments}
We performed an exhaustive assessment of the performance metrics of ternary-weight models derived through our fine-tuning methodology.
\begin{enumerate}
    \item conducted a comprehensive evaluation of the ternary-weight text-embedding model fine-tuned with our method.
    \item extended our proposed method to image-embedding models and evaluated its overall impact on real-word datasets.
    \item integrated our ternary-weight text-embedding model for retrieval tasks with Approximate Nearest Neighbor (ANN), evaluating the performance.
\end{enumerate}

\textbf{Configurations.} We utilize the xiaobu\footnote{\url{https://huggingface.co/lier007/xiaobu-embedding-v2}} text embedding model, which currently ranks as the state-of-the-art (SOTA) on the MTEB-Chinese leaderboard\footnote{\url{https://huggingface.co/spaces/mteb/leaderboard}}, and the ViT\footnote{\url{https://huggingface.co/google/vit-base-patch16-224}}~\cite{dosovitskiy:vit} image embedding model as benchmarks in our work. The original 32-bit full-precision linear layers of these models were replaced with ternary-weight ones prior to fine-tuning. The xiaobu model was fully fine-tuned on the nli-zh-25k\footnote{\url{https://huggingface.co/datasets/shibing624/nli-zh-all/tree/main/sampled_data}} and t2ranking\footnote{\url{https://huggingface.co/datasets/sentence-transformers/t2ranking/tree/main/triplet}} datasets, while the ViT model underwent full fine-tuning on the ImageNet-1k~\cite{deng:imagenet} training set. The label information is not used. Sentences or images are processed by both the full-precision and ternary-weight embedding models, with outputs from the full-precision model serving as targets for the ternary-weight model. Loss is then computed and backpropagated. The specific training parameters are detailed in Table~\ref{tbl:config}, where (1e-3, 2, 0.5) indicates that the initial learning rate is 1e-3 and is halved every 2 epochs. The default values apply to any unspecified parameters. The experimental equipment is an NVIDIA A800 GPU.

\begin{table*}[!htb]
\centering
\caption{Fine-tuning configurations for ternary-weight embedding models.}
\label{tbl:config}
\resizebox{0.5\textwidth}{!}{%
\begin{tabular}{c | cc}
\toprule
\toprule
&\textbf{Text Embedding}&\textbf{Image Embedding}\\
\midrule
backbone&xiaobu-embedding-v2&ViT-base\\
train set&nli-zh-25k + t2ranking& ImageNet-1k\\
loss&MSE&MSE\\
optimizer & Adam & Adam\\
epochs& 2 &  10 \\
learning rate & (2e-5, 1, 0.2)&  (1e-3, 2, 0.5)\\

\bottomrule
\bottomrule
\end{tabular}
}
\end{table*}

\textbf{Tasks and Metrics.} Massive Text Embedding Benchmark~\cite{muennighoff:mteb} (MTEB) is a massive benchmark for measuring the performance of text embedding models on diverse embedding tasks. We conducted a comprehensive analysis of the full-precision and ternary-weight text embedding models on all Chinese-language tasks, including retrieval, semantic textual similarity (STS), clustering, classification, reranking, and pairclassification. The reported results represent the main score for each task. In parallel, ImageNet-1k, CIFAR-10, and CIFAR-100~\cite{krizhevsky2009:cifar}, which are widely used datasets for evaluating visual models, were employed to train classifiers for full-precision and ternary-weight image embedding models. The reported results reflect classification accuracy on these datasets. Additionally, we assessed the forward speed and memory usage of each embedding model. For both the text and image embedding models, a single forward pass was performed on the nli-zh-25k dataset and the ImageNet-1k validation set, respectively, with the execution time recorded for comparison. When testing the latency and memory usage of ternary-weight models, we utilized the BITBLAS library\footnote{\url{https://github.com/microsoft/BitBLAS}}, which supports W2A16 computation.

\begin{table*}[!htb]
\centering
\caption{Main scores for the six MTEB-Chinese tasks of the full-precision model used as the baseline and the ternary-weight model fine-tuned from baseline using our method.}
\label{tbl:MTEB_xiaobu_TN_FP}
\resizebox{1\textwidth}{!}{%
\begin{tabular}{c | l l l l l l l l}
\toprule
\toprule
\textbf{Retrieval} & \textbf{Cmedqa} & \textbf{Covid} & \textbf{Du} & \textbf{Ecom} & \textbf{Medical} & \textbf{T2} & \textbf{Video} & \textbf{MMacro}\\
baseline & 47.18 & 89.45 & 89.45 & 70.57 & 68.13 & 85.01 & 80.16 & 82.27 \\
ours & 45.11 & 82.82 & 87.61 & 67.16 & 63.68 & 82.36 & 74.18 & 78.38 \\
\midrule

\textbf{STS} & \textbf{ATEC} & \textbf{BQ} & \textbf{LCQMC} & \textbf{PAWSX} & \textbf{QBQTC} & \textbf{STSB} & \textbf{AFQMC} & \\
baseline & 58.81 & 75.08 & 79.82 & 47.42 & 45.14 & 82.05 & 60.96 &  \\
ours & 57.15 & 74.28 & 79.58 & 46.26 & 41.89 & 81.47 & 58.25 &  \\
\midrule

\textbf{Clustering} & \textbf{CLS-P2P} & \textbf{CLS-S2S} & \textbf{ThuNews-P2P} & \textbf{ThuNews-S2S} &  &  &  & \\
baseline & 59.84 & 51.01 & 81.32 & 71.60 &  &  &  &  \\
ours & 54.68 & 48.54 & 77.21 & 67.40 &  &  &  &  \\
\midrule

\textbf{Classification} & \textbf{Waimai} & \textbf{Multilingual} & \textbf{JDReview} & \textbf{OnelineShopping} &  &  &  & \\
baseline & 88.34 & 77.79 & 87.62 & 94.15 &  &  &  &  \\
ours & 86.19 & 75.09 & 86.87 & 93.13 &  &  &  &  \\
\midrule

\textbf{ReRanking} & \textbf{CMedQAv1} & \textbf{CMedQAv2} & \textbf{MMarco} & \textbf{T2} &  &  &  & \\
baseline & 91.08 & 90.46 & 40.11 & 69.13 &  &  &  &  \\
ours & 88.74 & 88.38 & 35.04 & 68.38 &  &  &  &  \\
\midrule

\textbf{PairClassification} & \textbf{Cmnli} & \textbf{Ocnli} &  & &  &  &  & \\
baseline & 86.42 & 85.44 &  &  &  &  &  &  \\
ours & 85.14 & 83.22 &  &  &  &  &  &  \\
\bottomrule
\bottomrule
\end{tabular}
}
\end{table*}

\subsection{Ternary-Weight Text Embedding Model}
This section presents the evaluation results of our ternary-weight text embedding model. We evaluated its inference performance, latency, and storage usage relative to the full-precision baseline, assessed its performance and latency against the same-scale full-precision model, and compared our method against the currently popular quantization method, GPTQ.~\cite{frantar:gptq}. 

\textbf{Comparison with full-precision baseline.} We evaluated our ternary-weight text embedding model and its full-precision counterpart on MTEB-Chinese. In Table~\ref{tbl:MTEB_xiaobu_TN_FP}, the full-precision and ternary-weight models achieved average scores of 76.53/72.66, 64.18/62.70, 65.94/61.96, 72.70/70.14, and 85.93/84.18, respectively, across six types of tasks and In 19 out of 29 tasks, the score difference between full-precision  and ternary-weight models did not exceed 3. This level of performance loss is acceptable, as the values of all linear layers in the ternary-weight model were constrained to \{1, 0, -1\}, which reduced the model's inference latency and storage usage to 0.37$\times$ and 0.13$\times$ that of a 32-bit full-precision model, respectively, as shown in Table~\ref{tbl:text_embedding_speed_memory}. 

\begin{table*}[!htb]
\centering
\caption{Main scores for the six MTEB-Chinese tasks of our ternary-weight model and the full-precision counterpart model with comparable storage usage to our ternary-weight model.}
\label{tbl:MTEB_xiaobu_stella}
\resizebox{1\textwidth}{!}{%
\begin{tabular}{c | l l l l l l l l}
\toprule
\toprule
\textbf{Retrieval} & \textbf{Cmedqa} & \textbf{Covid} & \textbf{Du} & \textbf{Ecom} & \textbf{Medical} & \textbf{T2} & \textbf{Video} & \textbf{MMacro}\\
FP-counterpart & 42.05 & 79.78 & 86.49 & 62.84 & 58.50 & 82.20 & 68.29 & 78.34 \\
ours & \textbf{45.11} & \textbf{82.82} & \textbf{87.61} & \textbf{67.16} & \textbf{63.68} & \textbf{82.36} & \textbf{74.18} & \textbf{78.38} \\
\midrule

\textbf{STS} & \textbf{ATEC} & \textbf{BQ} & \textbf{LCQMC} & \textbf{PAWSX} & \textbf{QBQTC} & \textbf{STSB} & \textbf{AFQMC} & \\
FP-counterpart & 51.22 & 64.81 & 76.89 & 29.70 & 37.38 & 79.63 & 46.73 &  \\
ours & \textbf{57.15} & \textbf{74.28} & \textbf{79.58} & \textbf{46.26} & \textbf{41.89} & \textbf{81.47} & \textbf{58.25} &  \\
\midrule

\textbf{Clustering} & \textbf{CLS-P2P} & \textbf{CLS-S2S} & \textbf{ThuNews-P2P} & \textbf{ThuNews-S2S} &  &  &  & \\
FP-counterpart & 39.60 & 37.04 & 61.64 & 56.67 &  &  &  &  \\
ours & \textbf{54.68} & \textbf{48.54} & \textbf{77.21} & \textbf{67.40} &  &  &  &  \\
\midrule

\textbf{Classification} & \textbf{Waimai} & \textbf{Multilingual} & \textbf{JDReview} & \textbf{OnelineShopping} &  &  &  & \\
FP-counterpart & 85.50 & 70.52 & 84.44 & 90.77 &  &  &  &  \\
ours & \textbf{86.19} & \textbf{75.09} & \textbf{86.87} & \textbf{93.13} &  &  &  &  \\
\midrule

\textbf{ReRanking} & \textbf{CMedQAv1} & \textbf{CMedQAv2} & \textbf{MMarco} & \textbf{T2} &  &  &  & \\
FP-counterpart & 84.70 & 85.31 & 28.04 & 65.93 &  &  &  &  \\
ours & \textbf{88.74} & \textbf{88.38} & \textbf{35.04} & \textbf{68.38} &  &  &  &  \\
\midrule

\textbf{PairClassification} & \textbf{Cmnli} & \textbf{Ocnli} &  & &  &  &  & \\
FP-counterpart & 76.21 & 70.28 &  &  &  &  &  &  \\
ours & \textbf{85.14} & \textbf{83.22} &  &  &  &  &  &  \\
\bottomrule
\bottomrule
\end{tabular}
}

\end{table*}
\textbf{Comparison with same-scale model.} We chose the stella-base-zh\footnote{\url{https://huggingface.co/infgrad/stella-base-zh-v2}} model (which is denoted as FP-counterpart) as a reference, as it demonstrates the highest overall performance among models on the MTEB leaderboard while maintaining a storage usage comparable to our ternary-weight model. We evaluated the performance of both models and the results are summarized in Table~\ref{tbl:MTEB_xiaobu_stella}. With both models occupying similar storage usage, our ternary-weight outperforms the same-scale full-precision model across all tasks.

\begin{table*}[!htb]
\centering
\caption{Main scores for the six MTEB-Chinese tasks of the GPTQ-quantized full-precision baseline model and the ternary-weight models fine-tuned from baseline using our method.}
\label{tbl:MTEB_gptq}
\resizebox{1\textwidth}{!}{%
\begin{tabular}{c | l l l l l l l l}
\toprule
\toprule
\textbf{Retrieval} & \textbf{Cmedqa} & \textbf{Covid} & \textbf{Du} & \textbf{Ecom} & \textbf{Medical} & \textbf{T2} & \textbf{Vedio} & \textbf{MMacro}\\
GPTQ-g-1 & 34.49 & 62.68 & 66.71 & 49.71 & 45.21 & 65.11 & 56.06 & 51.13 \\
GPTQ-g16 & 44.37 & \textbf{84.07} & 87.11 & 65.64 & 62.02 & 81.87 & 73.36 & 76.59 \\
ours & \textbf{45.11} & 82.82 & \textbf{87.61} & \textbf{67.16} & \textbf{63.68} & \textbf{82.36} & \textbf{74.18} & \textbf{78.38} \\
\midrule

\textbf{STS} & \textbf{ATEC} & \textbf{BQ} & \textbf{LCQMC} & \textbf{PAWSX} & \textbf{QBQTC} & \textbf{STSB} & \textbf{AFQMC} & \\
GPTQ-g-1 & 43.68 & 57.53 & 72.76 & 11.69 & 26.44 & 73.28 & 37.20 &  \\
GPTQ-g16 & 55.76 & 70.38 & 78.53 & 34.51 & 37.68 & 79.38 & 56.18 &  \\
ours & \textbf{57.15} & \textbf{74.28} & \textbf{79.58} & \textbf{46.26} & \textbf{41.89} & \textbf{81.47} & \textbf{58.25} &  \\
\midrule

\textbf{Clustering} & \textbf{CLS-P2P} & \textbf{CLS-S2S} & \textbf{ThuNews-P2P} & \textbf{ThuNews-S2S} &  &  &  & \\
GPTQ-g-1 & 42.26 & 39.58 & 62.67 & 56.23 &  &  &  &  \\
GPTQ-g16 & 52.28 & 46.90 & 72.20 & \textbf{68.63} &  &  &  &  \\
ours & \textbf{54.68} & \textbf{48.54} & \textbf{77.21} & 67.40 &  &  &  &  \\
\midrule

\textbf{Classification} & \textbf{Waimai} & \textbf{Multilingual} & \textbf{JDReview} & \textbf{OnelineShopping} &  &  &  & \\
GPTQ-g-1 & 83.70 & 69.85 & 83.88 & 89.85 &  &  &  &  \\
GPTQ-g16 & \textbf{87.39} & \textbf{75.69} & \textbf{87.39} & \textbf{93.24} &  &  &  &  \\
ours & 86.19 & 75.09 & 86.87 & 93.13 &  &  &  &  \\
\midrule

\textbf{ReRanking} & \textbf{CMedQAv1} & \textbf{CMedQAv2} & \textbf{MMarco} & \textbf{T2} &  &  &  & \\
GPTQ-g-1 & 79.19 & 79.42 & 17.16 & 65.63 &  &  &  &  \\
GPTQ-g16 & 87.94 & 87.52 & 31.63 & 67.84 &  &  &  &  \\
ours & \textbf{88.74} & \textbf{88.38} & \textbf{35.04} & \textbf{68.38} &  &  &  &  \\
\midrule

\textbf{PairClassification} & \textbf{Cmnli} & \textbf{Ocnli} &  & &  &  &  & \\
GPTQ-g-1 & 67.67 & 62.75 &  &  &  &  &  &  \\
GPTQ-g16 & 80.71 & 79.32 &  &  &  &  &  &  \\
ours & \textbf{85.14} & \textbf{83.22} &  &  &  &  &  &  \\
\bottomrule
\bottomrule
\end{tabular}
}
\end{table*}

\textbf{Comparison with GPTQ.} To assess the performance of our fine-tuning method, we compared it with the widely recognized quantization method, GPTQ~\cite{frantar:gptq}. We quantized the weights of the full-precision baseline model to 2 bits using GPTQ, with weight values constrained to \{0, 1, 2, 3\}. Results were obtained for both ungrouped quantization (GPTQ-g-1) and grouped quantization with a group size of 16 (GPTQ-g16), using 1000 samples for quantization, as detailed in Table~\ref{tbl:MTEB_gptq}. Our method applied a single unified scaling parameter$\gamma$ across all weights. Although the model quantized with GPTQ-g-1 employs multiple scaling parameters, our ternary-weight model significantly outperforms it while achieving comparable inference latency. The performance of the model quantized with GPTQ-g16 underperforms on all tasks except classification, where it slightly outperforms our ternary-weight model. But overall, their performance across tasks is similar. In terms of storage, GPTQ-g16 requires $1.68\times$ (as shown in Table~\ref{tbl:text_embedding_speed_memory}) more space than our ternary-weight one. Furthermore, due to the smaller group size, GPTQ-g16 is expected to have higher inference latency than both GPTQ-g-1 and our ternary-weight model.

\begin{table*}[!htb]
\centering
\caption{Comparison of latency (one epoch on the nli-zh-25k dataset) and storage usage across text-embedding models.}
\label{tbl:text_embedding_speed_memory}
\resizebox{0.7\textwidth}{!}{%
\begin{tabular}{c | c c c c c}
\toprule
\toprule
&baseline & ours & FP-counterpart & GPTQ-g-1 & GPTQ-g16\\
\midrule
weight type&FP32&\{-1, 0, 1\}&FP32&\{0, 1, 2, 3\}&\{0, 1, 2, 3\}\\
Latency&135s&50s&39s&48s&-\\
Storage Usage&1241MB&\textbf159MB&197MB&231MB&267MB\\
\bottomrule
\bottomrule
\end{tabular}
}
\end{table*}

\textbf{Latency and Memory Usage.} We performed standardized latency and memory usage tests using the BITBLAS library on all the text embedding models mentioned above All, with the results summarized in Table~\ref{tbl:text_embedding_speed_memory}. In our experiments, the weights and inputs/outputs of full-precision models are in FP32 format. The weights of all the linear layers of our ternary-weight models are in INT2 format due to the absence of computational kernels specifically designed for the ternary values \{-1, 0, 1\} and the inputs/outputs are in FP16 format. The weights and input/output for the other layers are all in FP32 format. For GPTQ-g-1 and GPTQ-g16, quantization necessitates the storage of weights and zero points, leading both to use INT2 format. Consequently, the linear layers with merged weights and zero-points employ INT4 weights and FP16 input/output, while the weights and input/output for the remaining layers remain in FP32 format.

\subsection{Ternary-Weight Image Embedding Model}
This section assesses the effectiveness of our fine-tuning approach applied to image-embedding models. We fine-tuned the full-precision ViT model on the ImageNet-1k dataset to derive the ternary-weight one and incorporated classifiers into both models to evaluate the quality of the embeddings. The evaluation was conducted using the CIFAR10, CIFAR100, and ImageNet-1K datasets. The training parameters for the classifiers are detailed in Table~\ref{tbl:config_image_classification} and 
the default values apply to any unspecified parameters.

\begin{table*}[!htb]
\centering
\caption{Configurations of image classification tasks of imagenet embedding models.}
\label{tbl:config_image_classification}
\resizebox{0.45\textwidth}{!}{%
\begin{tabular}{c | ccc}
\toprule
\toprule
&cifar-10&cifar-100&ImageNet-1k\\
\midrule
optimizer & Adam & Adam & Adam\\
epochs& 10 & 10 & 1 \\
learning rate & 1e-3 & 1e-3 & 1e-3\\

\bottomrule
\bottomrule
\end{tabular}
}
\end{table*}

As shown in Table~\ref{tbl:image_embedding}, the validation performance gap between the full-precision and our ternary-weight image embedding models on the CIFAR10, CIFAR100, and ImageNet-1K datasets is -2.00$\%$, 4.02$\%$, and -1.57$\%$, respectively.In terms of inference speed, our ternary-weight model operates at 0.69$\times$ the speed of the full-precision model, while its storage usage is an impressive 0.07$\times$ that of the full-precision model.Notably, when we train only the classifier while keeping the embedding model frozen as the backbone, we achieve acceleration during the training phase as well, not just during inference.

\begin{table*}[!htb]
\centering
\caption{The classification accuracy of the full-precision baseline and our ternary-weight image embedding model on the CIFAR-10, CIFAR-100, and ImageNet datasets, along with their latency on the ImageNet validation set and storage uasge.}
\label{tbl:image_embedding}
\resizebox{0.8\textwidth}{!}{%
\begin{tabular}{c | c c c c c}
\toprule
\toprule
&\multicolumn{3}{c}{Accuracy ($\%$)}&\multirow{2}{*}{Latency}&\multirow{2}{*}{Storage Usage}\\
&CIFAR10&CIFAR100&ImagNet-1k&&\\
\midrule
baseline&96.79&86.38&76.35&122s&328M\\
ours&94.79&82.36&74.78&84s (0.69$\times$)&24M (0.07$\times$)\\
\bottomrule
\bottomrule
\end{tabular}
}
\end{table*}

\subsection{Integration of Ternary-Weight Text Embedding with ANN}
Online systems have a greater demand for model lightweightness. Therefore, we tested the performance of our ternary embedding model integrated into ANN methods using the popular FAISS library~\footnote{\url{https://github.com/facebookresearch/faiss}}. The dataset utilized is CmedqaRetrieval~\footnote{\url{https://huggingface.co/datasets/C-MTEB/CmedqaRetrieval}}. As illustrated in Table ~\ref{tbl:ANN-Embedding}, we compared the performance of our teranry-weight text-embedding model with that of the full-precision model across four ANN algorithms: FlatL2 (brute-force matching), IVFlat, LSH~\cite{datar:LSH}, and HNSW~\cite{malkov:HNSW}.The ternary-weight text embedding model obtained through our fine-tuning approach reduces the time needed to generate sentence embeddings to approximately $0.40\times\sim0.42\times$ that of the full-precision model. In terms of retrieval results, our ternary-weight model achieved comparable levels of precision and recall to the full-precision model, even surpassing it in some cases.

\begin{table*}[!htb]
\centering
\caption{Precision, recall and embedding generation time on CmedqaRetrieval dataset using the full-precision text embedding model and the ternary-weight text embedding model obtained with our method. Four ANN methods were used to evaluate the quality of embeddings.}
\label{tbl:ANN-Embedding}
\resizebox{1\textwidth}{!}{%
\begin{tabular}{c | c c c c c c c c}
\toprule
\toprule
\textbf{ANN} & \textbf{EmbeddingModel} & \textbf{Precision@1 ($\%$)} & \textbf{Precision@5 ($\%$)} &\textbf{Precision@10 ($\%$)} & \textbf{Recall@1 ($\%$)}&\textbf{Recall@5 ($\%$)}&\textbf{Recall@10 ($\%$)}&\textbf{Embedding Generating Time (s)}\\
\midrule
\multirow{2}{*}{Faiss: FlatL2} & xiaobu-FP &\textbf{40.61}&\textbf{16.37}& \textbf{9.93}  &\textbf{26.69}&\textbf{46.30}&\textbf{54.88}& 23.73\\
 & xiaobu-TN &40.26&16.17& 9.89 &26.53&46.00&54.58& \textbf{9.76 (0.41$\times$)}\\
 
 \midrule
\multirow{2}{*}{Faiss: IVFlat} & xiaobu-FP &31.08&11.81& 7.10 &20.07&33.16&39.23& 24.69\\
 & xiaobu-TN &\textbf{33.06}&\textbf{12.85}& \textbf{7.73} &\textbf{21.34}&\textbf{36.15}&\textbf{42.37}& \textbf{9.79 (0.40$\times$)}\\
 
 \midrule
\multirow{2}{*}{Faiss: LSH} & xiaobu-FP &\textbf{22.68}&\textbf{9.31}& 5.72 &\textbf{13.85}&\textbf{25.89}&31.44& 23.69\\
 & xiaobu-TN &21.73&9.17& \textbf{5.96} &13.35&25.56&\textbf{32.72}& \textbf{9.76 (0.42$\times$)}\\
 
 \midrule
\multirow{2}{*}{Faiss: HNSW} & xiaobu-FP &\textbf{39.73}&\textbf{16.06}& \textbf{9.76} &\textbf{26.03}&\textbf{45.22}&\textbf{53.65}& 24.45\\
 & xiaobu-TN &39.16&15.74& 9.64 &25.74&44.54&52.99& \textbf{9.72 (0.40$\times$)}\\
\bottomrule
\bottomrule
\end{tabular}
}
\end{table*}

\section{Discussion and Future Work}~\label{sec:discussion}

\textbf{Kernel for ternary-weight matrix computation.} The experiments regarding latency and storage usage presented in this paper were performed using the BITBLAS library, employing INT2 weights and FP16 inputs and outputs. Notably, there is an additional numerical redundancy, as INT2 values encompass -1, 0, 1, and 2, while ternary weights consist of -1, 0, and 1. For binary computers, achieving a memory compression of 2 bits per unit is already quite extreme. However, a more suitable ternary computing kernel could provide enhanced acceleration. Furthermore, our Eq~\ref{eq:forward} allows for controllable sparsity through the adjustment of $\beta$, as illustrated in Table~\ref{tbl:beta_sparsity}. Nonetheless, existing sparse computing kernels~\footnote{\url{https://github.com/NVIDIA/TensorRT}} do not support extremely low-bit computations like INT2, and our current method cannot produce a structurally sparse weight matrix.

\textbf{Optimal choice of $\beta$.} We bypass the limitation of weight distribution by introducing a predefined parameter $\beta$, though its selection is empirical. As shown in Table~\ref{tbl:MTEB_xiaobu_different_beta} of Appendix~\ref{apd:beta}, a value of 2 yields relatively good fine-tuning results. However, there is no theoretical guarantee for the optimal choice of $\beta$. An interesting observation is the positive correlation between $\beta$ and weight sparsity. When $\beta$ is set to 2, the weight sparsity rate is 0.583, which closely aligns with the optimal probability recommended for dropout. The relationship between $\beta$ and model performance warrants further exploration, both experimentally and theoretically.

\vspace{0.5 cm}
\bibliography{v0_draft}

\clearpage
\appendix
\noindent\textbf{\Large Appendix}\\~\\
This appendix provides the supplementary materials for this work, “Ternary Embedding Models” constructed according to the corresponding sections therein.

\section{Empirically Determined Partitioning Boundaries}\label{apd:beta}

There is currently no theoretically optimal solution for the threshold $\gamma$ in Eq~\eqref{eq:forward}. Ternary Neural Networks (TWN)~\cite{li:TWN} typically use 0.75 times the mean of the absolute weight values as a basis, but this approach relies on the assumption that weights follow a normal distribution--a condition that often does not hold in practice. Instead, we introduce $\beta$ to empirically seek an optimal value. Table~\ref{tbl:MTEB_xiaobu_different_beta} shows model performance across different values of $\beta$ when fine-tuned under identical parameters. Due to time and equipment constraints, we only tested three cases: $\beta=1$, $\beta=2$, and $\beta=3$, finding that 
$\beta=2$ yielded significantly better results than the other two. Additionally, as $\beta$ increases, the proportion of zero elements in the weights, or sparsity, also rises, as shown in Table~\ref{tbl:beta_sparsity}.

\begin{table*}[!htb]
\centering
\caption{The impact of $\beta$ on weight sparsity}
\label{tbl:beta_sparsity}
\resizebox{0.35\textwidth}{!}{%
\begin{tabular}{c | ccc}
\toprule
\toprule
$\beta$&1&2&3\\
\midrule
Sparsity&31.9\%&58.3\%&77.6\%\\
\bottomrule
\bottomrule
\end{tabular}
}
\end{table*}

A practical approach to model compression would involve empirically establishing the relationship between $\beta$, weight sparsity, and model performance to balance efficiency and effectiveness. However, the resources required for this are considerable. Moreover, TWN has shown that the optimal threshold lacks a closed-form solution, making it exceedingly difficult to theoretically determine the best $\beta$.

\begin{table*}[!htb]
\centering
\caption{The impact of $\beta$ on model performance}
\label{tbl:MTEB_xiaobu_different_beta}
\resizebox{1\textwidth}{!}{%
\begin{tabular}{c | l l l l l l l l}
\toprule
\toprule
\textbf{Retrieval} & \textbf{Cmedqa} & \textbf{Covid} & \textbf{Du} & \textbf{Ecom} & \textbf{Medical} & \textbf{T2} & \textbf{Video} & \textbf{MMacro}\\
$\beta=1$ & 42.16 & 76.71 & 85.81 & 63.51 & 60.38 & 79.88 & 70.68 & 75.47 \\
\textbf{$\beta=2$} & \textbf{45.11} & \textbf{82.82} & \textbf{87.61} & \textbf{67.16} & \textbf{63.68} & \textbf{82.36} & \textbf{74.18} & \textbf{78.38} \\
$\beta=3$ & 42.44 & 77.24 & 84.96 & 62.96 & 59.58 & 78.09 & 67.64 & 74.62 \\
\midrule

\textbf{STS} & \textbf{ATEC} & \textbf{BQ} & \textbf{LCQMC} & \textbf{PAWSX} & \textbf{QBQTC} & \textbf{STSB} & \textbf{AFQMC} & \\
$\beta=1$ & 56.04 & 73.80 & 79.38 & 44.85 & 37.52 & 81.11 & 56.04 &  \\
\textbf{$\beta=2$} & \textbf{57.15} & \textbf{74.28} & \textbf{79.58} & \textbf{46.26} & \textbf{41.89} & \textbf{81.47} & \textbf{58.25} &  \\
$\beta=3$ & 55.77 & 73.01 & 79.07 & 45.60 & 38.30 & 80.84 & 55.81 &  \\
\midrule

\textbf{Clustering} & \textbf{CLS-P2P} & \textbf{CLS-S2S} & \textbf{ThuNews-P2P} & \textbf{ThuNews-S2S} &  &  &  & \\
$\beta=1$ & 48.22 & 44.77 & 69.05 & 61.30 &  &  &  &  \\
\textbf{$\beta=2$} & \textbf{54.68} & \textbf{48.54} & \textbf{77.21} & \textbf{67.40} &  &  &  &  \\
$\beta=3$ & 51.42 & 45.41 & 72.00 & 63.02 &  &  &  &  \\
\midrule

\textbf{Classification} & \textbf{Waimai} & \textbf{Multilingual} & \textbf{JDReview} & \textbf{OnelineShopping} &  &  &  & \\
$\beta=1$ & 85.22 & 70.73 & 84.58 & 91.28 &  &  &  &  \\
\textbf{$\beta=2$} & \textbf{86.19} & \textbf{75.09} & \textbf{86.87} & \textbf{93.13} &  &  &  &  \\
$\beta=3$ & 85.48 & 73.29 & 84.50 & 92.70 &  &  &  &  \\
\midrule

\textbf{ReRanking} & \textbf{CMedQAv1} & \textbf{CMedQAv2} & \textbf{MMarco} & \textbf{T2} &  &  &  & \\
$\beta=1$ & 86.99 & 85.59 & 27.78 & 68.35 &  &  &  &  \\
\textbf{$\beta=2$} & \textbf{88.74} & \textbf{88.38} & \textbf{35.04} & \textbf{68.38} &  &  &  &  \\
$\beta=3$ & 87.09 & 86.31 & 31.97 & 67.59 &  &  &  &  \\
\midrule

\textbf{PairClassification} & \textbf{Cmnli} & \textbf{Ocnli} &  & &  &  &  & \\
$\beta=1$ & 82.95 & 76.77 &  &  &  &  &  &  \\
\textbf{$\beta=2$} & \textbf{85.14} & \textbf{83.22} &  &  &  &  &  &  \\
$\beta=3$ & 83.94 & 79.21 &  &  &  &  &  &  \\
\bottomrule
\bottomrule
\end{tabular}
}
\end{table*}

\section{Experiments on More Models}

To further validate the general effectiveness of our method, we applied the same fine-tuning process to the full-precision model stella-base-v2, maintaining the same training and testing setup as used in our experiments with xiaobu (see Table~\ref{tbl:config}). The performance results are shown in Table~\ref{tbl:MTEB_stella_FP_TN} and the comparison of model latency and storage requirements is provided in Table~\ref{tbl:stella_speed_memory}.

\begin{table*}[!htb]
\centering
\caption{Evaluation on MTEB-Chinese for stella-FP and stella-TN}
\label{tbl:MTEB_stella_FP_TN}
\resizebox{1\textwidth}{!}{%
\begin{tabular}{c | l l l l l l l l}
\toprule
\toprule
\textbf{Retrieval} & \textbf{Cmedqa} & \textbf{Covid} & \textbf{Du} & \textbf{Ecom} & \textbf{Medical} & \textbf{T2} & \textbf{Video} & \textbf{MMacro}\\
stella-FP & 42.05 & 79.78 & 86.49 & 62.84 & 58.50 & 82.20 & 68.29 & 78.34 \\
stella-TN & 37.78 & 72.15 & 82.22 & 57.85 & 53.57 & 75.10 & 61.09 &71.84 \\
\midrule

\textbf{STS} & \textbf{ATEC} & \textbf{BQ} & \textbf{LCQMC} & \textbf{PAWSX} & \textbf{QBQTC} & \textbf{STSB} & \textbf{AFQMC} & \\
stella-FP & 51.22 & 64.81 & 76.89 & 29.70 & 37.38 & 79.63 & 46.73 &  \\
stella-TN & 49.48 & 64.27 & 75.98 & 27.59 & 34.27 & 79.09 & 44.46 &  \\
\midrule

\textbf{Clustering} & \textbf{CLS-P2P} & \textbf{CLS-S2S} & \textbf{ThuNews-P2P} & \textbf{ThuNews-S2S} &  &  &  & \\
stella-FP & 39.60 & 37.04 & 61.64 & 56.67 &  &  &  &  \\
stella-TN & 38.77 & 36.35 & 59.84 & 53.01 &  &  &  &  \\
\midrule

\textbf{Classification} & \textbf{Waimai} & \textbf{Multilingual} & \textbf{JDReview} & \textbf{OnelineShopping} &  &  &  & \\
stella-FP & 85.50 & 70.52 & 84.44 & 90.77 &  &  &  &  \\
stella-TN & 82.92 & 66.06 & 82.91 & 88.54 &  &  &  &  \\
\midrule

\textbf{ReRanking} & \textbf{CMedQAv1} & \textbf{CMedQAv2} & \textbf{MMarco} & \textbf{T2} &  &  &  & \\
stella-FP & 84.70 & 85.31 & 28.04 & 65.93 &  &  &  &  \\
stella-TN & 81.21 & 81.90 & 21.72 & 64.81 &  &  &  &  \\
\midrule

\textbf{PairClassification} & \textbf{Cmnli} & \textbf{Ocnli} &  & &  &  &  & \\
stella-FP & 76.21 & 70.28 &  &  &  &  &  &  \\
stella-TN & 75.57 & 68.06 &  &  &  &  &  &  \\
\bottomrule
\bottomrule
\end{tabular}
}
\end{table*}

\begin{table*}[!htb]
\centering
\caption{Comparison of latency (one epoch on the nli-zh-25k dataset) and storage usage for full-precison baseline (stella-base-v2) and our ternary-weight model fine-tuned from baseline.}
\label{tbl:text_embedding_speed_memory}
\resizebox{0.35\textwidth}{!}{%
\begin{tabular}{c | c c}
\toprule
\toprule
&baseline & ours\\
\midrule
weight type&FP32&\{-1, 0, 1\}\\
Latency&39s&17s\\
Storage Usage&197MB&86MB\\
\bottomrule
\bottomrule
\end{tabular}
}
\end{table*}

\end{document}